\PassOptionsToPackage{table}{xcolor}
\documentclass[sigconf,screen]{acmart}

\copyrightyear{2026}
\acmYear{2026}
\setcopyright{cc}
\setcctype{by}
\acmConference[KDD '26]{Proceedings of the 32nd ACM SIGKDD Conference on Knowledge Discovery and Data Mining V.2}{August 09--13, 2026}{Jeju Island, Republic of Korea}
\acmBooktitle{Proceedings of the 32nd ACM SIGKDD Conference on Knowledge Discovery and Data Mining V.2 (KDD '26), August 09--13, 2026, Jeju Island, Republic of Korea}
\acmDOI{10.1145/3770855.3818888}
\acmISBN{979-8-4007-2259-2/2026/08}

\usepackage[utf8]{inputenc} %
\usepackage[T1]{fontenc}    %
\usepackage{hyperref}       %

\usepackage{url}            %
\usepackage{xurl}           %
\newcommand{\modelname}[1]{\path{#1}}
\usepackage{booktabs}       %
\usepackage{amsfonts}       %
\usepackage{nicefrac}       %
\usepackage{microtype}      %
\usepackage{multirow}
\usepackage{booktabs}
\usepackage{natbib}
\usepackage{tabularx}
\usepackage{graphicx} 
\usepackage{subcaption}
\usepackage{caption}
\usepackage{wrapfig}
\usepackage{enumitem}
\usepackage[most]{tcolorbox}
\usepackage{lipsum} %
\usepackage{cleveref}
\usepackage{bbding}
\usepackage{placeins}
\newcommand{\ours}{\textbf{MatSciBench}}

\begin{document}

\title[\ours: Benchmarking the Reasoning Ability of Large Language Models in Materials Science]{\ours: Benchmarking the Reasoning Ability of\\Large Language Models in Materials Science}

\author{Junkai Zhang}
\authornote{Equal contribution.}
\affiliation{%
  \institution{University of California, Los Angeles}
  \department{Computer Science Department}
  \city{Los Angeles}
  \state{California}
  \country{USA}
}
\email{zhang@cs.ucla.edu}

\author{Jingru Gan}
\authornotemark[1]
\affiliation{%
  \institution{University of California, Los Angeles}
  \department{Computer Science Department}
  \city{Los Angeles}
  \state{California}
  \country{USA}
}
\email{jrgan@cs.ucla.edu}

\author{Xiaoxuan Wang}
\affiliation{%
  \institution{University of California, Los Angeles}
  \department{Computer Science Department}
  \city{Los Angeles}
  \state{California}
  \country{USA}
}
\email{xw27@cs.ucla.edu}

\author{Zian Jia}
\affiliation{%
  \institution{University of Pennsylvania}
  \department{Department of Materials Science and Engineering}
  \city{Philadelphia}
  \state{Pennsylvania}
  \country{USA}}
\email{jiazian@foxmail.com}

\author{Changquan Gu}
\affiliation{%
  \institution{University of California, Los Angeles}
  \department{Computer Science Department}
  \city{Los Angeles}
  \state{California}
  \country{USA}
}
\email{zachkoo@cs.ucla.edu}

\author{Jianpeng Chen}
\affiliation{%
 \institution{Virginia Tech}
 \department{Department of Computer Science}
 \city{Blacksburg}
 \state{Virginia}
 \country{USA}}
\email{jianpengc@vt.edu}

\author{Yanqiao Zhu}
\affiliation{%
  \institution{University of California, Los Angeles}
  \department{Computer Science Department}
  \city{Los Angeles}
  \state{California}
  \country{USA}
}
\email{yzhu@cs.ucla.edu}

\author{Mingyu Derek Ma}
\affiliation{%
  \institution{University of California, Los Angeles}
  \department{Computer Science Department}
  \city{Los Angeles}
  \state{California}
  \country{USA}
}
\email{ma@derek.ma}

\author{Dawei Zhou}
\affiliation{%
 \institution{Virginia Tech}
 \department{Department of Computer Science}
 \city{Blacksburg}
 \state{Virginia}
 \country{USA}}
\email{zhoud@vt.edu}

\author{Ling Li}
\affiliation{%
  \institution{University of Pennsylvania}
  \department{Department of Materials Science and Engineering}
  \city{Philadelphia}
  \state{Pennsylvania}
  \country{USA}}
\email{lzli@seas.upenn.edu}

\author{Wei Wang}
\affiliation{%
  \institution{University of California, Los Angeles}
  \department{Computer Science Department}
  \city{Los Angeles}
  \state{California}
  \country{USA}
}
\email{weiwang@cs.ucla.edu}
\renewcommand{\shortauthors}{Zhang et al.}

\begin{abstract}
Large Language Models (LLMs) have shown strong scientific reasoning ability, but their performance on materials science problems remains less studied.
To fill this gap, we introduce \ours{}, a comprehensive college-level benchmark comprising 1,340 problems that span the essential subdisciplines of materials science.
\ours{} features a structured and fine-grained taxonomy that categorizes materials science questions into 6 primary fields and 31 subfields, together with a three-tier difficulty classification based on the reasoning length needed to solve each problem.
\ours{} includes detailed reference solutions for 946 questions, supports process-level error analysis, and contains 315 questions with images for evaluating multimodal reasoning.
We evaluate leading thinking and non-thinking LLMs on \ours{}, and further test three reasoning methods for non-thinking models: basic chain-of-thought prompting, tool augmentation, and self-correction.
The results show that current models still face clear limits in college-level materials science reasoning. \texttt{DeepSeek-R1} achieves the highest score on text-only questions at 75.22\% accuracy, and \texttt{GPT-5} performs the best on questions with images at 53.02\%.
Our analysis shows that tool augmentation improves many non-thinking models in a token-efficient way, while self-correction often fails to provide reliable gains and can revise correct answers into incorrect ones.
We further analyze performance across difficulty levels, reasoning efficiency, multimodal reasoning, and failure patterns, and find that current models are mainly limited by domain knowledge gaps, calculation errors, problem comprehension failures, and difficulty in extracting precise information from scientific figures.
We also evaluate small language models and find that they remain far behind larger models on knowledge-intensive materials science reasoning.
Overall, \ours{} provides a clear testbed for measuring current LLM limitations and guiding future work on scientific reasoning in materials science.\\
\raisebox{-0.2em}{\includegraphics[height=1em]{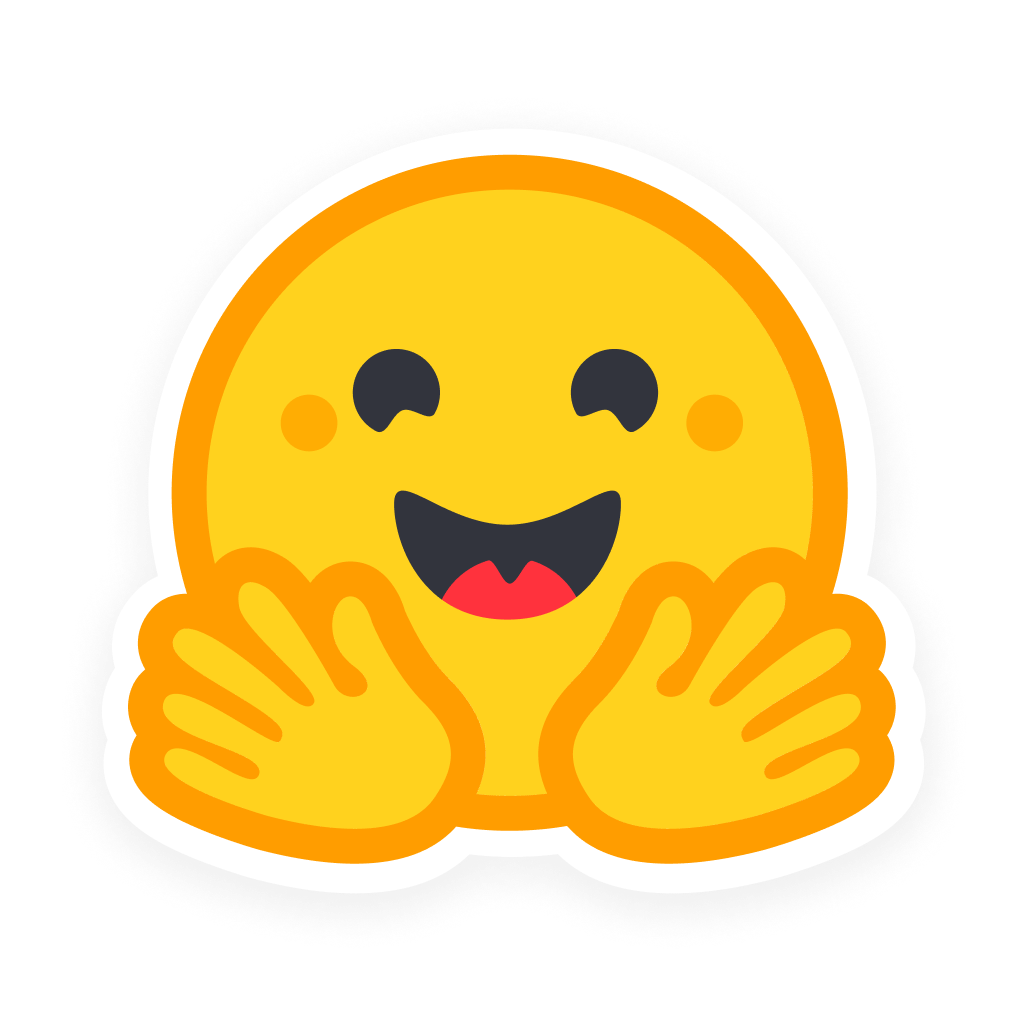}}~\textbf{Dataset:} \href{https://huggingface.co/datasets/JunkaiZ/MatSciBench}{huggingface.co/datasets/JunkaiZ/MatSciBench}\\
\raisebox{-0.2em}{\includegraphics[height=1em]{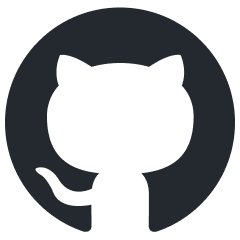}}~\textbf{Code:} \href{https://github.com/Jun-Kai-Zhang/MatSciBench.git}{github.com/Jun-Kai-Zhang/MatSciBench.git}
\end{abstract}



\maketitle

\section{Introduction}
Recent years have witnessed remarkable advancements of LLM reasoning abilities. From Chain of Thought (CoT) \citep{wei2022chain} to self-correction \citep{shinn2023reflexion} and tool augmentation \citep{gou2023tora}, the boundaries of LLM reasoning have expanded dramatically. What began with grade-school arithmetic calculations \citep{cobbe2021training} has evolved to solving problems at the level of International Mathematical Olympiad (IMO) silver medalists \citep{deepmind2024imo}. OpenAI’s o-series models can even solve a substantial portion of frontier mathematical problems that would typically require hours of concentrated effort from expert mathematicians \citep{openai2025o3mini,glazer2024frontiermath}.

Beyond LLMs' notable achievements in mathematics, general scientific reasoning has emerged as a new area of interest, where solving problems requires a proper combination of reasoning and domain-specific knowledge \citep{truhn2023large,ma2024clibench,ma2024sciagent}. Scientific reasoning benchmarks reveal that LLMs struggle to identify correct scientific assumptions and often demonstrate flawed understanding of scientific formulas and principles \citep{wang2023scibench}. Those findings indicate that scientific reasoning presents unique challenges to LLMs compared to pure mathematical questions. Therefore, numerous benchmarks have been proposed to assess LLMs' scientific reasoning capability, spanning from grade-school \citep{lu2022learn} to PhD-level \citep{feng2025physics} problems across domains \citep{huang2024chemeval,acharya2023nuclearqa}.

Despite the abundance of scientific problem-solving benchmarks, LLMs' reasoning abilities in materials science remain underexplored. Materials science occupies a unique position at the intersection of physics and chemistry, bridging fundamental science and engineering applications. This interdisciplinary field inherently relies on knowledge integration across multiple domains and requires complex reasoning capabilities. Existing reasoning benchmarks in materials science are limited by the lack of comprehensive evaluation and correct solutions \citep{zaki2024mascqa}, or by the dependence on synthetic data generated by LLMs themselves, which introduces unavoidable noise \citep{alamparamacbench}. In addition, none of the existing benchmarks adequately assesses the multimodal reasoning ability of LLMs in materials science.%

To comprehensively evaluate LLMs' reasoning abilities in materials science, we propose \ours{}, a benchmark comprising 1340 meticulously curated questions from 10 college-level textbooks spanning essential subdisciplines of materials science. All questions are open-ended to prevent model guessing while enabling objective assessment through rule-based judgment. For structured evaluation, \ours{} constructs a comprehensive and fine-grained taxonomy with 6 primary fields (Materials, Properties, Structures, Fundamental Mechanisms, Processes, Failure Mechanisms) and 31 subfields that capture materials science's interdisciplinary nature, enabling assessment of reasoning abilities on specific domains.
In addition, questions are classified into three difficulty levels based on reasoning length required to solve the question, with 57.2\% easy, 27.2\% medium, and 15.6\% hard questions. The 209 hard questions require long solving process, deliberately challenging models' complex reasoning capabilities. Detailed solutions to 946 of the questions are included to facilitate error categorization and process-level evaluations. The benchmark also incorporates 315 questions with images to assess multimodal reasoning abilities.

The o-series models from OpenAI, such as \modelname{o4-mini}, along with \modelname{Gemini-2.5-Pro}, \modelname{DeepSeek-R1}, \modelname{GPT-5}, \modelname{Claude-Sonnet-4}, and \modelname{Qwen3-235B-A22B}, represent a new class of LLMs that exhibit complex reasoning by generating extended intermediate outputs before producing final answers. These models are referred to as \textit{thinking models} or reasoning models, distinguishing them from standard LLMs like \modelname{GPT-4.1}, \modelname{Claude-Sonnet-3.7}, \modelname{DeepSeek-V3}, \modelname{Llama-4-Maverick}, and \modelname{Kimi-K2-Instruct}, classified as \textit{non-thinking models} \citep{chen2025seed1.5}. 
We conduct extensive experiments on \ours{} to evaluate and compare the reasoning capabilities of these six thinking models against five non-thinking models in materials science problem solving. In addition, we also evaluate the effectiveness of self-correction and tool augmentation (i.e., integration of Python code) on non-thinking models in addition to the basic CoT. Our results indicate that while \modelname{DeepSeek-R1} leads with approximately 75\% accuracy, the best-performing non-thinking model \modelname{Llama-4-Maverick} achieves a comparable 70\% with tool augmentation.

Our systematic analysis examines LLM reasoning capabilities through reasoning efficiency, multimodal reasoning, and failure patterns. The key findings from our analysis are as follows. First, higher accuracy often comes with substantially longer outputs, showing a clear efficiency-accuracy trade-off. Thinking models generally achieve stronger performance by producing much longer reasoning traces, while tool augmentation provides a more token-efficient way to improve many non-thinking models. Second, questions with images cause a large performance drop compared with text-only questions for the same multimodal LLMs, showing that visual reasoning remains a major challenge in materials science. Further error analysis shows that visual data extraction is the dominant bottleneck, accounting for more than half of substantive multimodal errors, especially on mechanical plots and property plots that require precise numerical reading. Third, our failure analysis shows that model errors are mainly caused by domain knowledge gaps, calculation inaccuracies, and problem understanding deficiencies. Tool augmentation consistently reduces numerical errors, but self-correction does not lead to reliable improvement; instead, it often introduces false critiques, unnecessary revisions, or incorrect repairs. These findings suggest that improving materials-science reasoning requires not only longer reasoning outputs, but also better grounding in domain knowledge, more reliable extraction from scientific figures, and stronger external verification mechanisms.

Our contributions are listed as follows:
\begin{itemize}[leftmargin=*]
    \item We introduce \ours, a comprehensive and challenging materials science reasoning benchmark comprising 1340 expert-curated questions from college-level textbooks across essential subdisciplines, featuring a structured taxonomy of 6 primary fields and 31 subfields, three-tier difficulty classification, detailed solutions for 946 questions, and 315 questions with visual contexts for multimodal reasoning evaluation.
    \item We benchmark state-of-the-art LLMs, including six thinking models and five non-thinking models. Additionally, we test the non-thinking models with three popular reasoning methods. This provides the most comprehensive evaluation and comparison of reasoning capabilities in materials science across different models and methods.
    \item We present a comprehensive multi-dimensional analysis of LLM reasoning capability on \ours{}, covering difficulty-stratified performance, reasoning efficiency, efficiency-accuracy trade-offs, multimodal reasoning, and failure patterns. We also evaluate small language models on \ours{} to assess their ability to handle knowledge-intensive materials-science reasoning. This detailed evaluation provides a clearer understanding of current model limitations and offers guidance for future improvements in scientific reasoning models.
\end{itemize}

\section{Related Work}
As LLMs continue to develop reasoning abilities, solving scientific problems is considered a fundamental dimension and has been the focus of numerous benchmarks. GSM8K \citep{cobbe2021training}, MATH \citep{hendrycks2021measuring}, along with a series of benchmarks \citep{mirzadeh2024gsm} evaluated the mathematical abilities of language models. With the emergence of multimodal LLMs, MathVista \citep{lu2023mathvista} further includes visual contexts to benchmark the multimodal reasoning abilities. With the growth of reasoning capabilities, competitive level questions like OlympiadBench \citep{he2024olympiadbench} and PutnamBench \citep{tsoukalas2024putnambench}, and advanced graduate-level math like Frontier Math \citep{glazer2024frontiermath} and HARDMATH \citep{fan2024hardmath} set new standards for reasoning models.

Beyond mathematics, natural science questions involve not only reasoning but also domain knowledge, thus incentivizing increased interest, particularly in chemistry, physics, and biology \citep{welbl2017crowdsourcing,lu2022learn,rein2024gpqa}. SciBench \citep{wang2023scibench}, MMMU \citep{yue2024mmmu}, MMMU-Pro \citep{yue2024mmmupro} cover college-level scientific problem solving, which requires both domain knowledge and sophisticated reasoning. 
OlympicArena \citep{huang2024olympicarena} contributes Olympiad-level, multimodal problems across seven scientific fields, and SuperGPQA \citep{du2025supergpqa} further expands coverage to 285 graduate-level disciplines.
Besides problem solving, SciEval~\citep{sun2024scieval} and SciKnowEval~\citep{feng2024sciknoweval} evaluate multi-level capabilities of LLM in scientific domain. In addition to those general natural scientific reasoning benchmarks, a series of works \citet{acharya2023nuclearqa,li2025atmossci} focus on specific domains. PhysReason \citep{zhang2025physreason}, PHYSICS \citep{feng2025physics}, and MM-PhyQA \citep{anand2024mm} specialize in physics questions; ChemEval \citep{huang2024chemeval} benchmarks chemistry abilities; \citet{meshram2024electrovizqa} for electronics.

\section{Dataset}

\subsection{Data Collection and Processing}
For our materials science benchmark dataset, we curated a collection of problems from textbooks across multiple subfields. We selected widely-adopted undergraduate and graduate textbooks that include both comprehensive references (such as ``The Science and Engineering of Materials'') and specialized resources focusing on specific domains (such as ``Electronic, Magnetic, and Optical Materials''). The choice of textbooks was guided and validated by materials science experts. We first identified the major subfields of materials science and then selected textbooks in these areas that provide exercise solutions and are accessible online. These sources collectively provide diverse problem types that cover the breadth of materials science concepts. 
The full list of textbooks is presented in \Cref{tab:data_textbooks}.

\begin{table}[h]
\centering
\caption{Textbooks Used for Question-answer Collection}
\label{tab:data_textbooks}
\begin{tabularx}{\linewidth}{@{}X c@{}}
\toprule
\textbf{Textbook} & \textbf{\# QAs} \\
\midrule
\rowcolor{gray!20} Introduction to Materials Science for Engineers~\citep{shackelford2015introduction} & 349 \\
The Science and Engineering of Materials~\citep{askeland2003science} & 287 \\
\rowcolor{gray!20} Materials Science and Engineering: An Introduction~\citep{callister2020materials} & 61 \\
Fundamentals of Materials Science and Engineering: An Integrated Approach~\citep{CallisterRethwisch2016Fundamentals} & 393 \\
\rowcolor{gray!20} Mechanical Behavior of Materials~\citep{hosford2010mechanical} & 83 \\
Electronic, Magnetic, and Optical Materials~\citep{fulay2016electronic} & 72 \\
\rowcolor{gray!20} Materials and Process Selection for Engineering Design~\citep{farag2020materials} & 27 \\
Fundamentals of Ceramics~\citep{barsoum2019fundamentals} & 29 \\
\rowcolor{gray!20} Physical Metallurgy~\citep{hosford2010physical} & 27 \\
Polymer Science and Technology~\citep{fried2014polymer} & 12\\
\midrule
\textbf{Total} & \textbf{1340} \\
\bottomrule
\end{tabularx}
\rowcolors{1}{}{}
\end{table}

We used Mistral optical character recognition (OCR) \citep{mistralocr} to digitize both textual and visual content of these textbooks. Then we implemented a parsing algorithm to identify the example problems and solutions from the digital copies. Each question-answer pair was structured into a standardized format. Following the initial extraction, each entry was reviewed and corrected by humans with LLM assistance to ensure accuracy and completeness. We applied strict filtering criteria, retaining only questions with determinate answers in the form of numerical values or formulas.

\subsection{Dataset Statistics}
Our benchmark comprises 1340 question-answer pairs structured in a standardized format. Questions are categorized as either numerical or formula-derivation type according to the type field, with 1236 NUM questions (92.2\%) and 104 FORMULA questions (7.8\%). In addition, 315 questions (23.5\%) include images.

\subsection{Taxonomy Classification}

\begin{figure}[h]
\centering
\includegraphics[width=\linewidth]{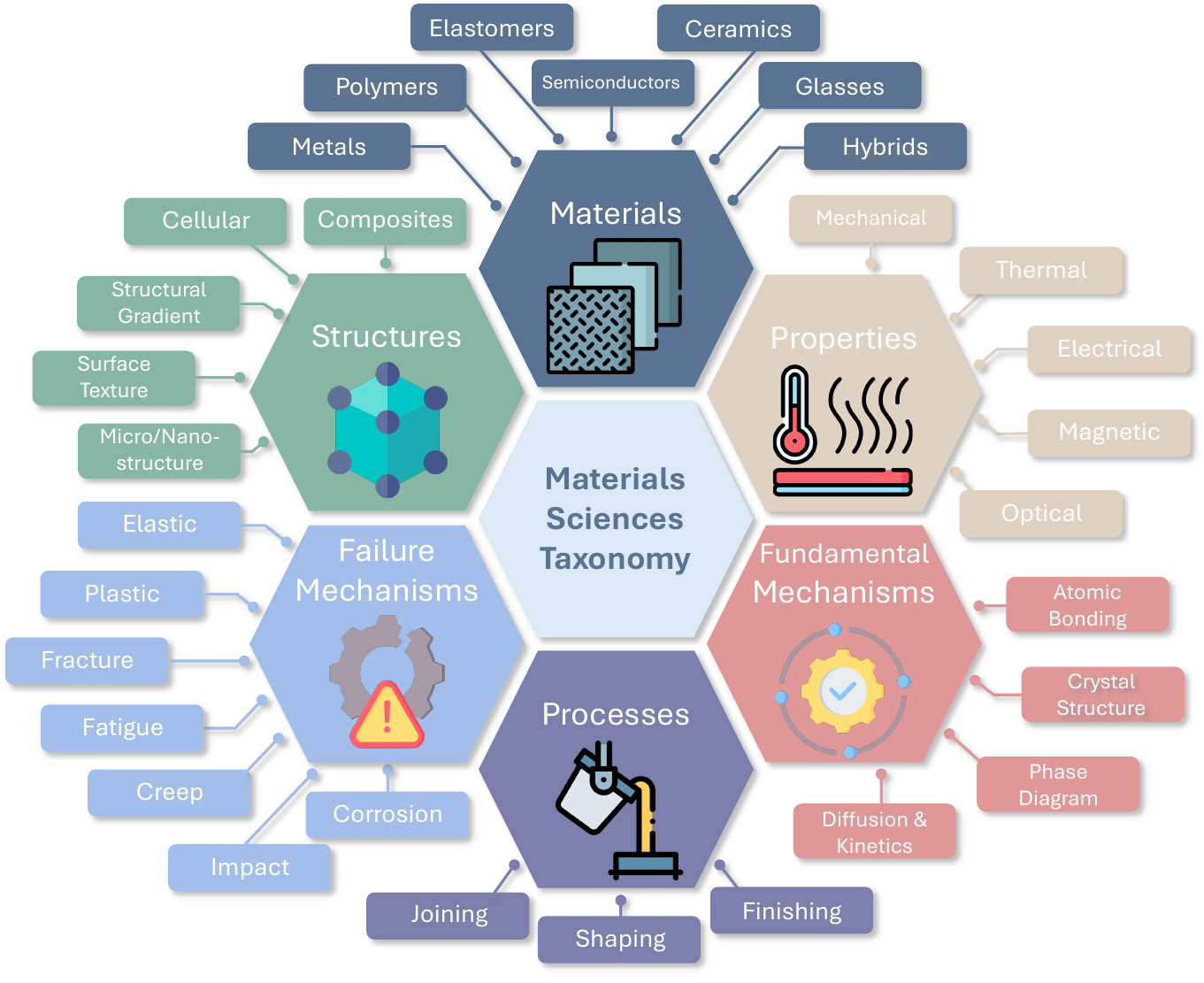}
\caption{Taxonomy of \ours{} Materials Science QAs.}
\label{fig:taxonomy}
\end{figure}

We developed a comprehensive hierarchical taxonomy to systematically categorize questions across fundamental materials science domains. Our taxonomy design was informed by established materials science curricula and reference texts, including \cite{shackelford2015introduction, shackelford2016crc, ashby2019materials}. The taxonomy framework follows the traditional organization of materials science education while covering key domains that characterize materials science problems, including materials, structures, properties, mechanisms, processes, and failure behaviors. The taxonomy consists of six primary fields, each containing detailed subcategories:

\begin{enumerate}[leftmargin=*, itemsep=1pt, parsep=0pt]
    \item \textbf{Materials}: Metals, Polymers, Elastomers, Semiconductors, Ceramics, Glasses, Hybrids
    \item \textbf{Properties}: Mechanical, Thermal, Electrical, Magnetic, Optical
    \item \textbf{Structures}: Composites, Cellular, Structural Gradient, Surface Texture, Micro/Nano-structure
    \item \textbf{Fundamental Mechanisms}: Atomic Bonding, Crystal Structure, Phase Diagram, Diffusion \& Kinetics
    \item \textbf{Processes}: Joining, Shaping, Finishing
    \item \textbf{Failure Mechanisms}: Elastic, Plastic, Fracture, Fatigue, Creep, Impact, Corrosion
\end{enumerate}

\Cref{fig:taxonomy} shows our manually developed taxonomy that covers the domains of QAs collected in \ours. The taxonomy consists of six primary fields: Materials, Properties, Structures, Fundamental Mechanisms, Processes, and Failure Mechanisms, each containing detailed subcategories. This multi-dimensional classification scheme enables us to capture the interdisciplinary nature of materials science problems, where a single question might span multiple domains. Our taxonomy not only provides a nuanced understanding of the dataset composition but also enables targeted evaluation of model performance across specific subfields and their intersections, offering insights into how AI models handle different aspects of materials science knowledge.

\subsection{Difficulty Classification}

We implemented a three-tier classification from easy, medium to hard, to assess question difficulty.
Following the previous work~\citet{guha2025openthoughtsdatarecipesreasoning}, difficulty is assessed with response lengths from LLMs. To further mitigate the bias of classification from a single model, we choose a collective of 6 models: \modelname{Claude-Sonnet-3.7}, \modelname{Claude-Sonnet-4}, \modelname{DeepSeek-V3}, \modelname{GPT-4.1}, \modelname{Llama-4-Maverick}, and \modelname{Qwen3-235B-A22B}, among which each pairwise Spearman rank correlation is above $0.5$. Each model's response length assigns a difficulty label to each question, and then we use majority voting to decide the final difficulty label. This gives us 766 easy questions, 365 medium questions, and 209 hard questions. This distribution provides coverage across difficulty levels, while differentiating questions in terms of the knowledge and reasoning length required to derive a correct solution.

\subsection{Data Leakage Detection}

Data leakage is a key concern when constructing reliable benchmarks. We use the method proposed in~\cite{xu2024benchmarking} to detect potential data leakage in our benchmark. Because this method is restricted to locally served models and the released implementation is not optimized for very large models, we evaluate a set of small language models: \modelname{Qwen-2.5-7B}, \modelname{Qwen-2.5-32B}, \modelname{Gemma-3-4B}, \modelname{Gemma-3-12B}, and \modelname{Gemma-3-27B}. The results, reported in \Cref{tab:data_leakage}, show N-gram accuracy and PPL only slightly above 1 for all models, indicating minimal overlap with training data. We found no evidence of leakage in the tested small models.

\begin{table}[h]
\centering
\caption{Data Leakage Detection on Small Language Models.}
\label{tab:data_leakage}
\begin{tabular}{lcc}
\toprule
\textbf{Model} & \textbf{N-gram Accuracy} & \textbf{PPL} \\
\midrule
\textbf{Qwen-2.5-Instruct-7B}  & 1.28 & 1.07 \\
\textbf{Qwen-2.5-Instruct-32B} & 1.22 & 1.06 \\
\textbf{Gemma-3-4B-it}         & 1.04 & 1.08 \\
\textbf{Gemma-3-12B-it}        & 1.06 & 1.10 \\
\textbf{Gemma-3-27B-it}        & 1.09 & 1.07 \\
\bottomrule
\end{tabular}
\end{table}

  \begin{tcolorbox}[colback=green!5!white, colframe=green!60!black, title=An Example Question from \ours{}, breakable]
  \textbf{Question}

  A steel alloy to be used for a spring application must have a modulus of resilience of at least $2.07 \, \text{MPa}$
  $(300 \, \text{psi})$. If the Young's modulus of the steel is $207 \, \text{GPa}$, what minimum yield strength must it
  have?

  \textbf{Image}\quad None

  \textbf{Solution}

  The modulus of resilience, yield strength, and elastic modulus are related by
  $$
  U_r = \frac{\sigma_y^2}{2E}.
  $$
  Solving for the yield strength gives
  $$
  \sigma_y = \sqrt{2 U_r E}.
  $$
  Using $U_r = 2.07 \, \text{MPa}$ and $E = 207 \times 10^3 \, \text{MPa}$,
  $$
  \sigma_y = \sqrt{(2)(2.07 \, \text{MPa})(207 \times 10^3 \, \text{MPa})}
  = 926 \, \text{MPa}.
  $$

  \textbf{Answer}\quad 926

  \textbf{Unit}\quad MPa

  \textbf{Primary Category}\quad Materials: Metals

  \textbf{Categories}

  Materials: Metals \\
  Properties: Mechanical \\
  Failure Mechanisms: Elastic

  \textbf{Type}\quad NUM

  \textbf{Difficulty Level}\quad Easy

  \textbf{Source Book}\quad Fundamentals of Materials Science and Engineering: An Integrated Approach
  \end{tcolorbox}

\section{Experiments}
\subsection{Models and Methods}
For proprietary models, we evaluate \modelname{GPT-4.1}~\citep{gpt41site}, \modelname{Claude-Sonnet-3.7}~\citep{anthropic2025claude37}, and the thinking models \modelname{o4-mini} \citep{openai2024o4mini}, \modelname{Gemini-2.5-Pro} \citep{deepmind2025geminipro}, \modelname{GPT-5}~\citep{openai2025gpt5}, \modelname{Claude-Sonnet-4}~\citep{anthropic2025claude}; for open-weight models, we evaluate \modelname{DeepSeek-V3}~\citep{liu2024deepseek}, \modelname{Kimi-K2-Instruct}~\citep{kimiteam2026kimik2openagentic}, \modelname{Llama-4-Maverick}~\citep{meta2025llama4}, and the thinking models \modelname{DeepSeek-R1} \citep{guo2025deepseek}, \modelname{Qwen3-235B-A22B}~\citep{yang2025qwen3}. Among these models, \modelname{GPT-4.1}, \modelname{Claude-Sonnet-3.7}, \modelname{o4-mini}, \modelname{Gemini-2.5-Pro}, \modelname{GPT-5}, \modelname{Claude-Sonnet-4}, and \modelname{Llama-4-Maverick} support visual inputs.

For non-thinking models, we adopt three prompting methods: basic CoT, self-correction, and tool augmentation. 
The self-correction method follows \citet{huang2023large,kim2023language,shinn2023reflexion}, invoking 3 rounds of conversation with the model: (1) the initial response, (2) detecting issues in the initial attempt, and (3) revising the initial attempt based on the detected problem. 
The tool augmentation method prompts the model to generate Python code, executes it using a code interpreter \citep{gou2023tora, yang2024qwen2}, and derives the final answer based on the execution results.
The detailed prompts are provided in \Cref{app:prompts}.

\begin{table*}[]
\centering
\small
\renewcommand{\arraystretch}{1.2}
\setlength{\tabcolsep}{5pt}
\caption{Experimental Results in Terms of Accuracy Score (\%) on \ours{} (questions without images). \textbf{Boldface} denotes the best result, and \underline{underlining} denotes the second-best result.}
\label{tab:text-only-results}
\begin{tabular}{lccccccc}
\toprule
\textbf{Model} & \textbf{Fail. Mech.} & \textbf{Fund. Mech.} & \textbf{Materials} & \textbf{Processes} & \textbf{Properties} & \textbf{Structures} & \textbf{Overall} \\
\midrule
\multicolumn{8}{c}{\textit{Non-Thinking Models}} \\
\textbf{Claude-Sonnet-3.7} & 62.64 & 58.40 & 62.14 & 55.56 & 60.95 & 63.06 & 62.54 \\
\quad\textit{+Correction} & 61.51 & 55.25 & 61.04 & 53.54 & 59.94 & 61.64 & 61.27 \\
\quad\textit{+Tool} & 66.79 & 59.03 & 64.46 & 56.57 & 64.27 & 64.92 & 64.78 \\
\textbf{DeepSeek-V3} & 65.28 & 63.66 & 67.55 & 60.61 & 65.27 & 67.21 & 67.02 \\
\quad\textit{+Correction} & 66.04 & 59.03 & 63.02 & 60.61 & 62.97 & 63.61 & 62.83 \\
\quad\textit{+Tool} & \underline{70.57} & 64.71 & 69.32 & \textbf{67.68} & 67.44 & 70.16 & 68.78 \\
\textbf{GPT-4.1} & 66.04 & 65.34 & 69.09 & 57.58 & 67.00 & 69.29 & 68.49 \\
\quad\textit{+Correction} & 64.53 & 61.76 & 67.77 & 53.54 & 65.99 & 68.42 & 67.32 \\
\quad\textit{+Tool} & 67.17 & \textbf{66.39} & 69.09 & 58.59 & \underline{68.16} & 69.51 & 68.39 \\
\textbf{Kimi-K2-Instruct} & \textbf{71.70} & \underline{65.97} & \underline{70.20} & \textbf{67.68} & 67.29 & \underline{70.49} & 68.78 \\
\quad\textit{+Correction} & \underline{70.57} & 63.87 & 67.44 & 64.65 & 65.99 & 68.20 & 66.93 \\
\quad\textit{+Tool} & 69.81 & 65.76 & 69.65 & 64.65 & 67.58 & 70.38 & \underline{69.46} \\
\textbf{Llama-4-Maverick} & 68.30 & 61.97 & 68.43 & 65.66 & 67.29 & 68.31 & 67.41 \\
\quad\textit{+Correction} & 67.92 & 63.03 & 68.76 & \underline{66.67} & 67.58 & 68.63 & 67.71 \\
\quad\textit{+Tool} & 70.19 & 65.76 & \textbf{71.41} & \underline{66.67} & \textbf{70.03} & \textbf{71.26} & \textbf{70.15} \\
\midrule
\multicolumn{8}{c}{\textit{Thinking Models}} \\
\textbf{Claude-Sonnet-4} & 71.32 & 64.08 & 69.09 & 65.66 & 67.29 & 68.96 & 68.49 \\
\textbf{DeepSeek-R1} & \underline{75.47} & \textbf{75.21} & \underline{75.83} & \underline{71.72} & \textbf{74.50} & \underline{76.17} & \textbf{75.22} \\
\textbf{Gemini-2.5-Pro} & \textbf{76.23} & 73.11 & 74.50 & 65.66 & 72.62 & 75.08 & 74.44 \\
\textbf{Qwen3-235B-A22B} & \underline{75.47} & 71.85 & 74.06 & 66.67 & 72.19 & 74.54 & 73.37 \\
\textbf{GPT-5} & 74.34 & \underline{74.58} & \textbf{76.05} & \textbf{74.75} & \underline{72.77} & \textbf{76.61} & \underline{75.02} \\
\textbf{o4-mini} & 71.32 & 69.12 & 71.96 & 68.69 & 70.89 & 72.02 & 71.22 \\
\bottomrule
\end{tabular}
\end{table*}

\begin{table*}[]
\centering
\small
\renewcommand{\arraystretch}{1.2}
\setlength{\tabcolsep}{5pt}
\caption{Experimental Results in Terms of Accuracy Score (\%) on \ours{} (questions with images). \textbf{Boldface} denotes the best result, and \underline{underlining} denotes the second-best result.}
\label{tab:multimodal-results}
\begin{tabular}{lccccccc}
\toprule
\textbf{Model} & \textbf{Fail. Mech.} & \textbf{Fund. Mech.} & \textbf{Materials} & \textbf{Processes} & \textbf{Properties} & \textbf{Structures} & \textbf{Overall} \\
\midrule
\textbf{Claude-Sonnet-3.7} & 28.57 & 47.31 & 42.54 & 43.75 & 36.84 & 42.25 & 40.95 \\
\textbf{Claude-Sonnet-4} & 35.71 & 47.90 & 47.01 & 53.13 & 42.58 & 46.13 & 44.44 \\
\textbf{Gemini-2.5-Pro} & \underline{46.43} & \underline{52.10} & \underline{54.10} & 56.25 & \underline{51.67} & \underline{52.46} & \underline{51.43} \\
\textbf{GPT-5} & \textbf{48.81} & \textbf{56.89} & \textbf{55.22} & \textbf{65.63} & \textbf{52.63} & \textbf{53.87} & \textbf{53.02} \\
\textbf{o4-mini} & 33.33 & 47.31 & 45.15 & \underline{59.38} & 42.58 & 44.72 & 44.44 \\
\textbf{GPT-4.1} & 25.00 & 46.71 & 41.42 & 43.75 & 34.45 & 41.55 & 40.00 \\
\textbf{Llama-4-Maverick} & 40.48 & 47.90 & 48.13 & 40.63 & 42.11 & 46.83 & 46.03 \\
\bottomrule
\end{tabular}
\end{table*}

\subsection{Evaluation}
The correctness of the output answers is evaluated using a hybrid approach that combines rule-based evaluation and LLM-based evaluation. We adapt the rule-based evaluation system from Qwen-2.5 Math \citep{yang2024qwen2}. Following the previous works~\citep{methani2020plotqa,gupta2024enhancing}, we apply a relaxed numerical tolerance of 5\% to account for approximation errors in calculations and image recognition. To address the limitations of rule-based systems in handling complex formulas and equations, we supplement this approach with an optional LLM judge for formula-type questions. The LLM's judgment serves as the final determinant of correctness for these complex mathematical expressions in our reported results. The performance in terms of accuracy score of all models on text-only questions is presented in \Cref{tab:text-only-results}, and the performance of multimodal models on questions with images is presented in \Cref{tab:multimodal-results}.

\subsection{Results}

\textbf{Observation 1.} Among the non-thinking models, \modelname{Llama-4-Maverick} with tool augmentation achieves the best overall accuracy (70.15\%). \modelname{Kimi-K2-Instruct} with tool augmentation obtains the second-best overall score (69.46\%), followed by \modelname{DeepSeek-V3} with tool augmentation and \modelname{Kimi-K2-Instruct} under basic CoT, both at 68.78\%. Under the basic CoT setting, \modelname{Kimi-K2-Instruct} performs best among non-thinking models (68.78\%), slightly ahead of \modelname{GPT-4.1} (68.49\%) and \modelname{Llama-4-Maverick} (67.41\%).

For thinking models, \modelname{DeepSeek-R1} achieves the best overall result (75.22\%), narrowly outperforming \modelname{GPT-5} (75.02\%). \modelname{Gemini-2.5-Pro} ranks third with 74.44\%, followed by \modelname{Qwen3-235B-A22B} (73.37\%) and \modelname{o4-mini} (71.22\%). \modelname{Claude-Sonnet-4} reaches 68.49\%, which is close to the strongest basic CoT non-thinking models but below the best tool-augmented non-thinking model. These results show that open-weight thinking models remain highly competitive on \ours{}, with \modelname{DeepSeek-R1} slightly exceeding \modelname{GPT-5} on text-only questions.

\textbf{Observation 2.} Tool augmentation improves the overall performance for most non-thinking models, whereas self-correction degrades the performance for 4 out of 5 models. Specifically, tool augmentation improves the overall accuracy of \modelname{Claude-Sonnet-3.7} from 62.54\% to 64.78\%, \modelname{DeepSeek-V3} from 67.02\% to 68.78\%, \modelname{Kimi-K2-Instruct} from 68.78\% to 69.46\%, and \modelname{Llama-4-Maverick} from 67.41\% to 70.15\%. For \modelname{GPT-4.1}, tool augmentation leads to only a negligible decrease from 68.49\% to 68.39\%, indicating that its performance remains largely unchanged. In contrast, self-correction reduces the overall accuracy for nearly all models, including \modelname{Claude-Sonnet-3.7} from 62.54\% to 61.27\%, \modelname{DeepSeek-V3} from 67.02\% to 62.83\%, \modelname{GPT-4.1} from 68.49\% to 67.32\%, and \modelname{Kimi-K2-Instruct} from 68.78\% to 66.93\%. The only exception is \modelname{Llama-4-Maverick}, whose accuracy slightly increases from 67.41\% to 67.71\%. This suggests that self-correction does not provide reliable gains for non-thinking models and can hurt model performance by revising originally correct answers.

\textbf{Observation 3.} In the multimodal evaluation, \modelname{GPT-5} delivers the strongest performance, achieving the highest overall accuracy (53.02\%) and leading across all individual categories. \modelname{Gemini-2.5-Pro} ranks second overall with 51.43\%, while \modelname{Llama-4-Maverick} ranks third with 46.03\%. \modelname{Claude-Sonnet-4} and \modelname{o4-mini} obtain the same overall accuracy (44.44\%), with \modelname{o4-mini} showing especially strong performance on the Processes category (59.38\%). \modelname{Claude-Sonnet-3.7} reaches 40.95\%, slightly ahead of \modelname{GPT-4.1} (40.00\%). These results show that \modelname{GPT-5} is the strongest model for multimodal materials science reasoning, while \modelname{Gemini-2.5-Pro} remains the closest competitor. However, the open-weight models still lag behind the leading proprietary models, suggesting that a clear gap remains between open-weight and proprietary models in multimodal materials science reasoning.

\subsection{Performance of Small Language Models on \ours{}}
Small language models (SLMs) have recently become promising for agentic AI~\citep{belcak2025small}. To assess their performance in the science domain, we evaluate the \modelname{Qwen-2.5} family~\citep{qwen2025qwen25technicalreport} and the \modelname{Gemma-3} family~\citep{gemmateam2025gemma3technicalreport} on \ours{}; the results are reported in \Cref{tab:small-models}. Overall, SLMs still struggle with knowledge-intensive scientific reasoning: even the best-performing model, \modelname{Qwen-2.5-Instruct-32b}, achieves only 44.88\% accuracy. Performance is strongly correlated with model size.

\begin{table*}[htbp]
\centering
\small
\renewcommand{\arraystretch}{1.2}
\setlength{\tabcolsep}{5pt}
\caption{Experimental Results in Terms of Accuracy Score (\%) on \ours{} (questions without images) for Small Language Models. \textbf{Boldface} denotes the best result, and \underline{underlining} denotes the second-best result.}
\label{tab:small-models}
\begin{tabular}{lccccccc}
\toprule
\textbf{Model} & \textbf{Fail. Mech.} & \textbf{Fund. Mech.} & \textbf{Materials} & \textbf{Processes} & \textbf{Properties} & \textbf{Structures} & \textbf{Overall} \\
\midrule
\textbf{Gemma-3-4b-it} & 9.81 & 9.24 & 12.36 & 7.07 & 10.95 & 12.46 & 12.39 \\
\textbf{Gemma-3-12b-it} & 21.13 & 13.87 & 22.85 & 13.13 & 21.47 & 23.17 & 22.93 \\
\textbf{Gemma-3-27b-it} & \underline{38.87} & \underline{31.30} & \underline{39.74} & \underline{30.30} & \underline{38.18} & \underline{39.45} & \underline{38.83} \\
\textbf{Qwen2.5-7b-instruct} & 21.51 & 22.48 & 26.49 & 20.20 & 24.64 & 27.10 & 26.83 \\
\textbf{Qwen2.5-32b-instruct} & \textbf{44.91} & \textbf{37.61} & \textbf{44.48} & \textbf{35.35} & \textbf{43.23} & \textbf{45.25} & \textbf{44.88} \\
\bottomrule
\end{tabular}
\end{table*}

\section{Analysis}

\subsection{Performance across Difficulty Levels}
\begin{figure*}[h]
\centering
\includegraphics[width=1.0\linewidth]{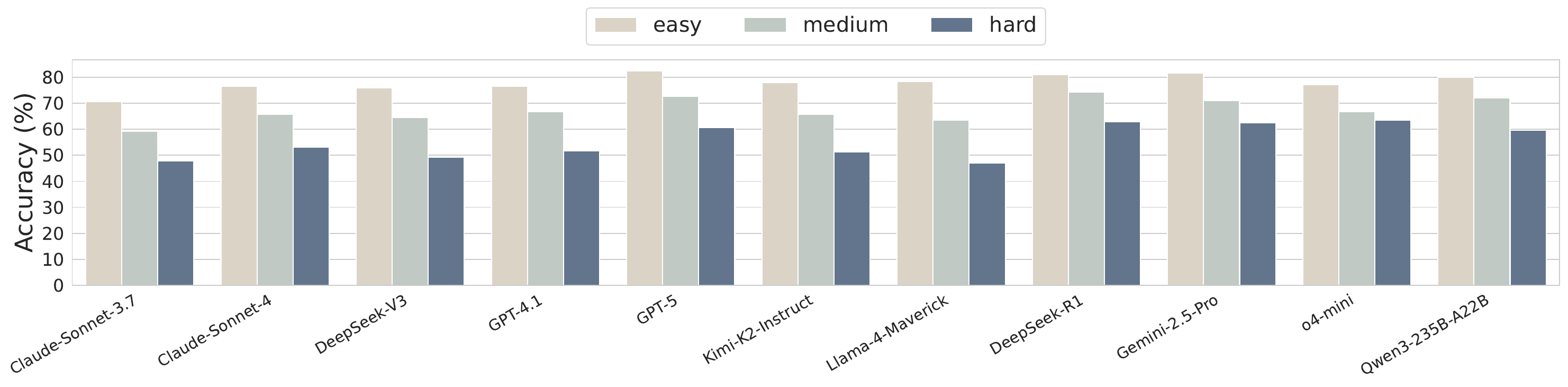}
\caption{The Performance of LLMs across Difficulty Levels. }
\label{fig:difficulty_level}
\end{figure*}

Our benchmark effectively differentiates models' scientific reasoning capabilities through progressively challenging tasks that require domain knowledge, calculation precision, and multi-step reasoning. 
The accuracy scores of different models across difficulty levels are shown in \Cref{fig:difficulty_level}.
Thinking models surpass non-thinking models by a significant margin at the most challenging level. This divergence highlights how reasoning capabilities become increasingly critical as problem complexity grows. 
All models exhibit performance degradation patterns with increasing difficulty, suggesting that complex reasoning processes prevent them from reaching correct answers.

\subsection{Efficiency versus Accuracy}

\begin{figure}[h]
\centering
\includegraphics[width=\linewidth]{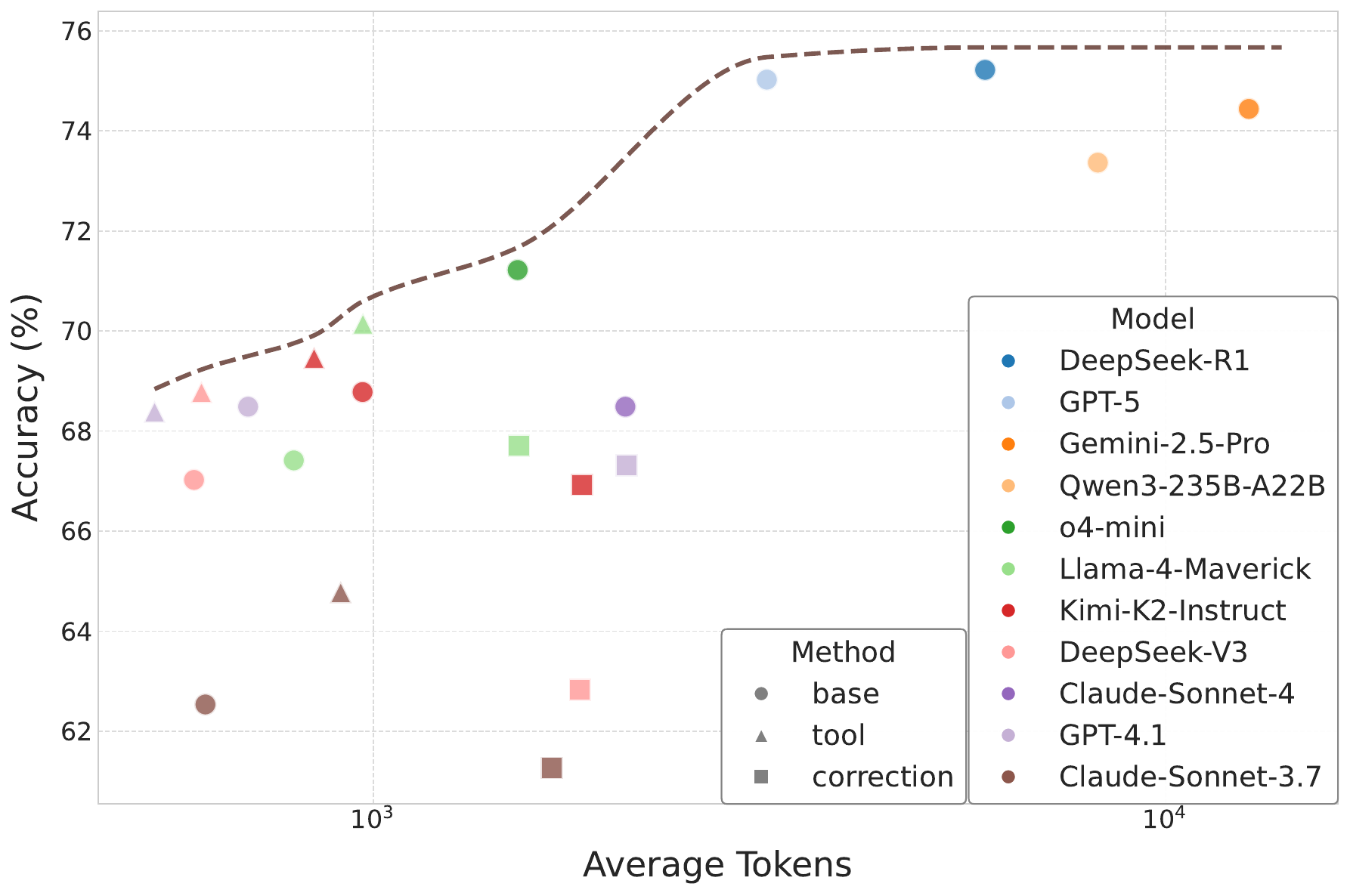}
\caption{Efficiency versus Accuracy.}
\label{fig:token_vs_accuracy}
\end{figure}

Thinking models often produce long outputs that include branching, backtracking, error checking, and correction \citep{yeo2025demystifying}. These behaviors can help models reach correct answers, but they also increase inference cost. This reveals a clear trade-off between reasoning accuracy and efficiency. \Cref{fig:token_vs_accuracy} shows this trade-off by comparing accuracy and average token usage across models and prompting methods, where the boundary line marks the reasoning efficiency frontier. Under basic CoT prompting, thinking models achieve higher accuracy than non-thinking models, but require substantially more tokens. Self-correction further increases output length, yet does not lead to consistent performance gains and can even reduce accuracy for some models. In contrast, tool augmentation offers a more token-efficient strategy: it adds only a small number of tokens while improving accuracy for many non-thinking models. Therefore, tool-augmented results often lie on the Pareto frontier, showing a better balance between accuracy and token cost.

\begin{figure*}[!h]
\centering
\includegraphics[width=1.0\linewidth]{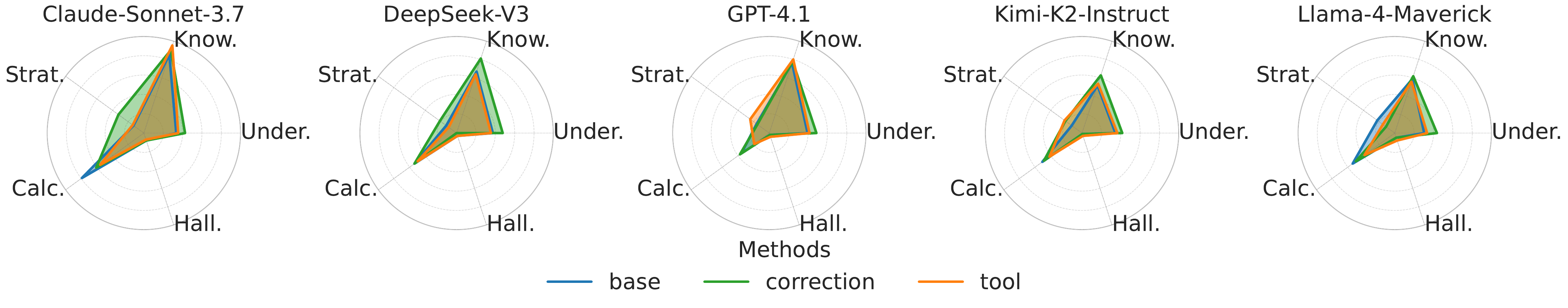}
\caption{Error categorization for non-thinking models. Error types are problem understanding deficiencies (under.), domain knowledge gaps (know.), flawed solution strategies (strat.), calculation inaccuracies (calc.), and hallucinated content (hall.).
}
\label{fig:error_categorization}
\end{figure*}

\subsection{Performance Drop Due to Visual Context}

\begin{figure}[h]
\centering
\includegraphics[width=\linewidth]{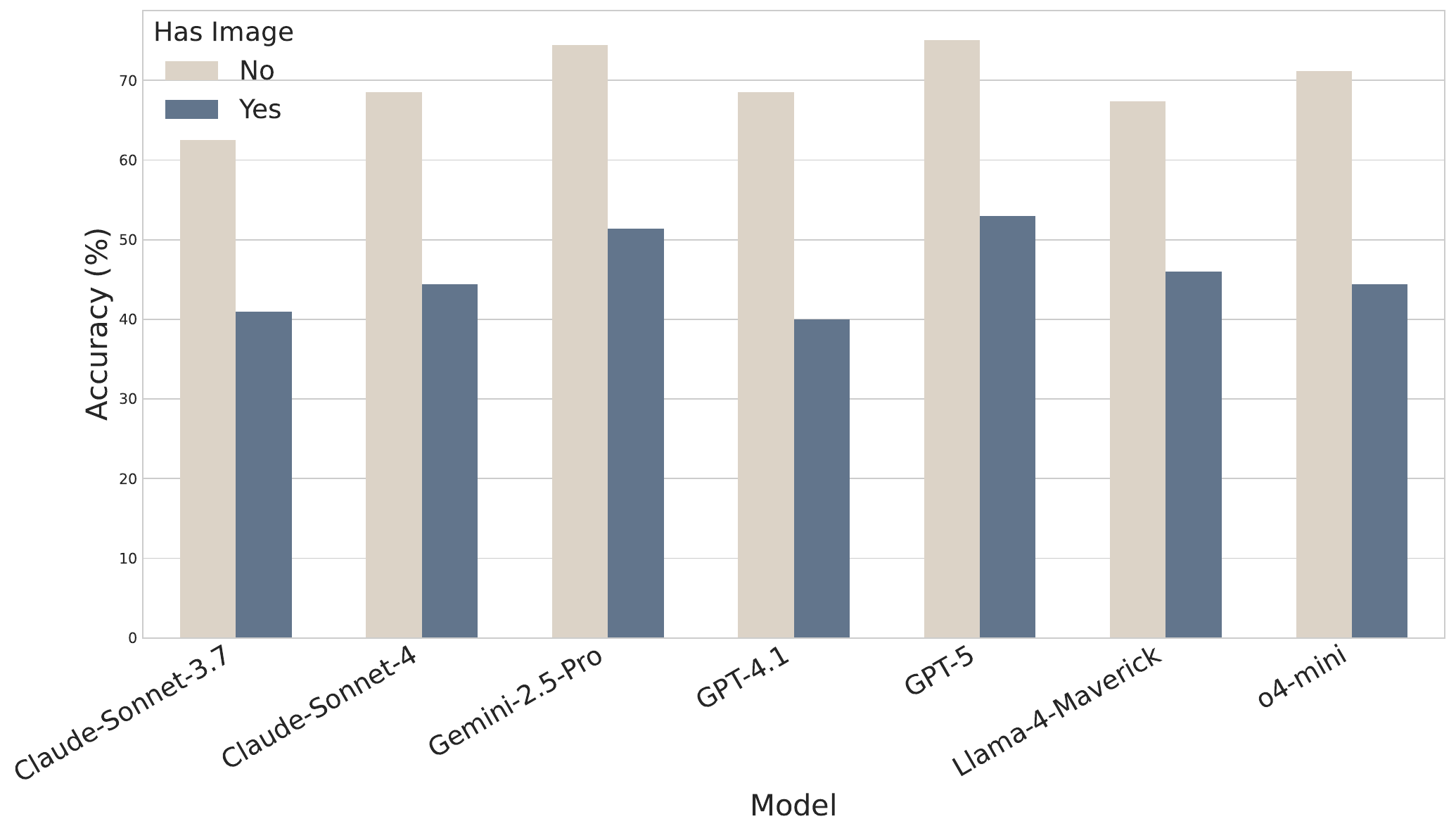}
\caption{Performance comparison of MLLMs on questions with and without images.}
\label{fig:image_vs_nonimage}
\end{figure}

Questions with images are significantly more challenging than text-only questions for multimodal LLMs, with significantly lower accuracy scores, as presented in \Cref{fig:image_vs_nonimage}. 

To further analyze the multimodal performance drop, we classify errors on questions with images into five substantive categories, excluding ambiguous, unparsed, and formatting-only cases. Across 781 substantive multimodal errors, Visual Data Extraction is the dominant error source, accounting for 68.9\%, with a range of 62.6\%--78.0\% across models. The other errors are less frequent: Domain Knowledge (16.9\%), Calculation Error (6.9\%), Spatial and Geometric Reasoning (4.1\%), and Problem Comprehension (3.2\%). This consistent pattern suggests that extracting correct information from scientific figures is a common bottleneck for multimodal LLMs.

We group images into eight types and calculate accuracy for types with more than 100 question-model pairs, excluding Micrograph due to its small sample size. The lowest accuracies occur on Mechanical Plot (26.8\%) and Property Plot (31.1\%), followed by  Phase Diagram (46.4\%), Crystal Structure (49.6\%), Mixed Visual (54.8\%), Engineering Schematic (64.6\%), and Data Table (66.5\%). The two hardest reported image types all require precise numerical reading from plots or diagrams. Cross-analysis further shows that Phase Diagram and Property Plot have the highest Visual Data Extraction shares among substantive errors, at 85.1\% and 85.0\%, respectively, mainly due to misread temperatures, axis values, or interpolation along curved boundaries.

\subsection{Failure Pattern}
To investigate the challenges LLMs face in solving materials science problems, we manually examined incorrect responses and identified five major error categories: problem understanding deficiencies, domain knowledge gaps, flawed solution strategies, calculation inaccuracies, and hallucinated content. To conduct systematic analysis of these error patterns automatically, we use an LLM to categorize mistakes across these five categories, evaluating all non-thinking models and prompting methods on text-only questions with reference solutions. When multiple errors exist, we classify them into the first appearing one in the solution.

The error rates across categories are presented in Figure~\ref{fig:error_categorization}. These findings reveal consistent patterns across all models, with deficiencies in domain knowledge, calculation errors, and problem understanding deficiencies the most critical limitations. While errors caused by hallucinations are still present, they occur less frequently than other error types. As expected, tool augmentation methods reduced numerical errors across all models. Self-correction methods, on the other hand, did not provide consistent improvements across any of the tested models in any error category.

\subsection{The Vulnerability of Self-correction}

We further analyze how answers change within standalone self-correction runs, comparing each model's initial answer to its final revised answer. Across the five non-thinking models, self-correction worsens 342 answers and improves 233, giving a net accuracy drop of 2.1\%. For initially correct answers that become incorrect, the main cause is false problem identification (69.6\%), where the model incorrectly critiques a correct original answer and revises it into a wrong one. The second most common cause is over-correction in rounding or precision (8.7\%), followed by introduced calculation errors (6.9\%), knowledge confusion (6.0\%), hallucinated constraints (5.4\%), and answer-extraction errors (3.3\%).

For initially wrong answers that remain wrong, the major failure mode is simply repeating the same answer (56.1\%), followed by producing a different wrong answer (28.1\%) and identifying the wrong problem in the review step (9.8\%). In contrast, when self-correction succeeds, it mainly does so by catching formula errors (28.6\%), calculation errors (25.1\%), conceptual errors (24.2\%), reading errors (15.2\%), or answer extraction errors (3.5\%). By difficulty, self-correction degrades easy questions by 2.6\%, medium questions by 1.8\%, and hard questions by 1.1\%. This decreasing degradation suggests that self-correction is most damaging when the model's initial answer is already likely to be correct, while on harder questions there are fewer correct answers to corrupt and more real errors that can potentially be fixed.

Overall, these results indicate that self-correction fails mainly because it depends on reliable self-evaluation. In materials science reasoning, checking whether a solution is correct often requires nearly the same domain knowledge and numerical care as solving the problem itself. Without an external verifier or grounded feedback, the review step can introduce false critiques, unnecessary changes, or incorrect repairs.

\section{Conclusion}

In this work, we present \ours{}, a college-level materials science reasoning benchmark containing 1,340 curated questions across core subdisciplines of materials science. We evaluate state-of-the-art thinking and non-thinking LLMs on \ours{}, and further test three reasoning methods for non-thinking models: basic chain-of-thought prompting, self-correction, and tool augmentation. The results show clear performance gaps across models and reveal that different reasoning methods have uneven effects: tool augmentation improves most non-thinking models, while self-correction often fails to provide reliable gains. We also evaluate small language models and find that they remain limited on knowledge-intensive scientific reasoning tasks. Beyond overall accuracy, we analyze model behavior across difficulty levels, reasoning efficiency, multimodal reasoning, and failure patterns. These analyses show that current models are constrained by difficulty in accurately extracting information from scientific figures, domain knowledge gaps, calculation errors, and problem comprehension failures. Overall, \ours{} provides a structured testbed for understanding current limitations and guiding future improvements in materials science reasoning.

\FloatBarrier

\section*{Limitations and Ethical Considerations}

\ours{} is designed to be broad and fine-grained, but it cannot fully represent the entire landscape of materials science curricula. Question distribution may contain residual subjectivity, and some topics may be under-represented. Our evaluation results are also sensitive to prompting, decoding choices, and model versions; moreover, closed-source models limit transparency regarding training data. There are no ethical concerns about this work.

\section*{Acknowledgements}
This work was partially supported by DARPA HR001126CE054, NIH OT2OD038003, U54OD036472, U54HG012517, U54DK097771, NSF 2312501, 2106859, Amazon, Optum, NEC.

\bibliographystyle{ACM-Reference-Format}
\bibliography{main}

\onecolumn
\appendix
\section{Experimental Details}\label{app:exp}
\subsection{Prompts}\label{app:prompts}
Prompts we used for each method are as follows. For every example, the original problem statement is sent as the user message. When a question has a non-empty unit field, we append the unit information to this user message before querying the model. For single-answer questions, we append ``The unit of the answer is \{unit\}.'' For multiple-answer questions, we append ``The units of each required answer are \{unit\}, respectively.'' The system prompts still instruct the model to place only the final answer inside \verb|\boxed{...}| and to omit the unit from the boxed answer.
  
  The Basic System Prompt is used in basic CoT and self-correction.
\begin{tcolorbox}[
    colframe=black,         %
    colback=white,          %
    coltitle=white,         %
    colbacktitle=black,     %
    fonttitle=\bfseries,    %
    title=Basic System Prompt,          %
    boxrule=1pt             %
]
You are a renowned materials science engineering professor with extensive knowledge in the field. Your students have presented you with a challenging question related to materials science. Please reason step by step, and put the final answer inside a single box using \verb|\boxed{...}|. Include only the final answer inside the box, without the unit.
\end{tcolorbox}

The tool augmentation is prompted to use Python code to improve the computation.
\begin{tcolorbox}[
    colframe=black,         %
    colback=white,          %
    coltitle=white,         %
    colbacktitle=black,     %
    fonttitle=\bfseries,    %
    title=Tool System Prompt,          %
    boxrule=1pt             %
]
You are a renowned materials science engineering professor with extensive knowledge in the field. Your students have presented you with a challenging question related to materials science. If necessary, you could write a single clean Python code block that computes necessary numeric values. Enclose the code in triple backticks with \verb|```python|. Please reason step by step, if no code is needed, put the final answer inside a single box using \verb|\boxed{...}|; otherwise, wait for the user to execute the code and give you the execution result, and then put the final answer inside a single box using \verb|\boxed{...}|. Include only the final answer inside the box, without the unit.
\end{tcolorbox}
After code execution, the model gets the results and uses the results to produce the final answer.
\begin{tcolorbox}[
    colframe=black,         %
    colback=white,          %
    coltitle=white,         %
    colbacktitle=black,     %
    fonttitle=\bfseries,    %
    title=Tool Summary Prompt,          %
    boxrule=1pt             %
]
Here are the results of the code execution:\textbackslash n\textbackslash n\{code\_executed\}\textbackslash n\textbackslash nBased on these results, what is the final answer to the original question?
\end{tcolorbox}

When using the self-correction, the model is first prompted to review and find the problem from its initial response.
\begin{tcolorbox}[
    colframe=black,         %
    colback=white,          %
    coltitle=white,         %
    colbacktitle=black,     %
    fonttitle=\bfseries,    %
    title=Review Prompt,          %
    boxrule=1pt             %
]
Review your previous answer and find problems with your answer.
\end{tcolorbox}
Then, the model is prompted to improve the initial response with the problem it found.
\begin{tcolorbox}[
    colframe=black,         %
    colback=white,          %
    coltitle=white,         %
    colbacktitle=black,     %
    fonttitle=\bfseries,    %
    title=Revise Prompt,          %
    boxrule=1pt             %
]
Based on the problems you found, improve your answer. Please reiterate your answer, with your final answer in the form \verb|\boxed{answer}|.
\end{tcolorbox}

\end{document}